# DeepQA: Improving the estimation of single protein model quality with deep belief networks

Renzhi Cao[1], Debswapna Bhattacharya[1], Jie Hou[1], and Jianlin Cheng[1, 2, §]

[1]Department of Computer Science, University of Missouri, Columbia, MO 65211, USA

[2]Informatics Institute, University of Missouri, Columbia, Missouri, 65211, USA

[§]Corresponding author (email: chengji@missouri.edu; address: Department of Computer Science, University of Missouri, Columbia, MO 65211, USA)

Emails:

RC: rcrg4@mail.missouri.edu

DB: db279@mail.missouri.edu

JH: jh7x3@mail.missouri.edu

JC: chengji@missouri.edu

## Abstract

## Background

Protein quality assessment (QA) by ranking and selecting protein models has long been viewed as one of the major challenges for protein tertiary structure prediction. Especially, estimating the quality of a single protein model, which is important for selecting a few good models out of a large model pool consisting of mostly low-quality models, is still a largely unsolved problem.

## Results

We introduce a novel single-model quality assessment method DeepQA based on deep belief network that utilizes a number of selected features describing the quality of a model from different perspectives, such as energy, physio-chemical characteristics, and structural information. The deep belief network is trained on several large datasets consisting of models from the Critical Assessment of Protein Structure Prediction (CASP) experiments, several publicly available datasets, and models generated by our in-house *ab initio* method. Our experiment demonstrate that deep belief network has better performance compared to Support Vector Machines and Neural Networks on the protein model quality assessment problem, and our method DeepQA achieves the state-of-the-art performance on CASP11 dataset. It also outperformed two well-established methods in selecting good outlier models from a large set of models of mostly low quality generated by *ab initio* modeling methods.

## Conclusion

DeepQA is a useful tool for protein single model quality assessment and protein structure prediction. The source code, executable, document and training/test datasets of DeepQA for Linux is freely available to non-commercial users at http://cactus.rnet.missouri.edu/DeepQA/.

# Introduction

The tertiary structures of proteins are important for understanding their functions, and have a lot of biomedical applications, such as the drug discovery [1]. With the wide application of next generation sequencing technologies, millions of protein sequences have been generated, which create a huge gap between the number of protein sequences and the number of protein structures [2, 3]. The computational structure prediction methods have the potential to fill the gap, since it is much faster and cheaper than experimental techniques, and also can be used for proteins whose structures are hard to be determined by experimental techniques, such as X-ray crystallography [1].

There are generally two major challenges in protein structure prediction [4]. The first challenge is how to sample the protein structural model from the protein sequences, the so-called structure sampling problem. Two different kinds of methods have been used to do the model sampling. The first is template-based modeling method [5-11] which uses the known structure information of homologous proteins as templates to build protein structure model, such as I-TASSER [12], FALCON [10, 11], MUFOLD [13] and RaptorX [14]. The second is *ab initio* modeling method [15-20], which builds the structure from scratch, without using existing template structure information. The second challenge is how to select good models from generated models pool, the so-called model ranking problem. It is essential for protein structure prediction, such as selecting models generated by *ab initio* modeling methods. There are mainly two different types of methods for the model ranking. The first is consensus methods [21-23], which calculate the average structural similarity score of a model against other models as its model quality. This method assumes the models in a model pool that are more similar to other models have better quality. It shows good performance in previous Critical Assessment of Techniques for Protein

Structure Prediction (CASP) experiments, which is a worldwide experiment for blindly testing protein structure prediction methods every two year. However, the accuracy of this method depends on input data, such as the proportion of good models in a model pool and the similarity between low quality models. It has been shown that this kind of methods is not working well when a large portion of models are of low quality [24]. The time complexity of most consensus methods is $O(n^2)$ time complexity (n: the total number of models), making it too slow to assess the quality of a large number of models. These problems with consensus methods highlight the importance of developing another kind of protein model quality assessment (QA) method – single-model QA method [5, 18, 24-30] that predicts the model quality based on the information from a single model itself. Single-model quality assessment methods only require the information of a single model as input, and therefore its performance does not depend on the distribution of high and low quality models in a model pool. In this paper, we focus on develop a new single-model quality assessment method that uses deep learning in conjunction with a number of useful features relevant to protein model quality.

Currently, most single-model QA methods predict the model quality from sequence evolutionary information [31], residue environment compatibility [32], structural features and physics-based knowledge [26-29, 33-35]. On such single-model QA method - ProQ2 [36] has relatively good performance in the CASP11 experiment, which uses Support Vector Machines with a number of features from a model and its sequence to predict its quality.

Here, we propose to develop a novel single-model quality assessment method based on deep belief network - a kind of deep learning methods that show a lot of promises in image processing [37-39] and bioinformatics [40]. We benchmark the performance of this method on large QA datasets, including the CASP datasets, four datasets from the recently 3DRobot decoys [41], and a dataset generated by our in-house *ab initio* modeling method UniCon3D. The good performance of our method - DeepQA on these datasets demonstrate the potential of applying deep learning techniques for protein model quality assessment.

The paper is organized as follows. In the Methods Section, we describe the datasets and features that are used for deep learning method, and how we implement, train, and evaluate the performance of our method. In the Result Section, we compare the performance of deep learning technique with two other QA methods based on support vector machines and neural networks. In the Results and Discussion Section, we summarize the results. In the Conclusion Section, we conclude the paper with our findings and future works.

## Methods

### Datasets

We collect three previous CASP models (CASP8, CASP9, and CASP10) from the CASP website [http://predictioncenter.org/download_area/](http://predictioncenter.org/download_area/), 3DRobot decoys[41], and 3113 native protein structure from PISCES database [42] as the training datasets. CASP11 models as testing dataset, and UniCon3D *ab initio* CASP11 decoys as the validation datasets. The 3DRobot decoys have four sets: 200 non-homologous (48 α, 40 β, and 112 α/β) single domain proteins each having 300 structural decoys; 58 proteins generated in a Rosetta benchmark[43] each having 100 structural decoys; 20 proteins in a Modeller benchmark [44] each having 200 structural decoys; and 56 proteins in a I-TASSER benchmark [16] each having 400 structural decoys. Two sets (stage1 and stage2) of CASP11 targets are used to test the performance of DeepQA. Each target at stage 1 contains 20 server models spanning the whole range of structural quality and each target at stage 2 contains 150 top server models selected by Davis-QAconsensus method. In total, 803 proteins with 216,875 structural decoys covering wide range of qualities are collected for training and testing DeepQA. All of these data and calculated quality scores are available at: http://cactus.rnet.missouri.edu/DeepQA/. In addition, we validate performance of our QA methods in a dataset produced by our *ab initio* modeling tool UniCon3D, which in total includes 24 targets and 20030 models.

**Input features for DeepQA**

In total, 16 features are used for benchmarking our method DeepQA, which describe the structural, physio-chemical and energy properties of a protein model. These features include 9 available top-performing energy and knowledge-based potentials scores, including ModelEvaluator score [28], Dope score [29], RWplus score [27], RF_CB_SRS_OD score [26], Qprob scores[30], GOAP score [45], OPUS score [46], ProQ2 score [36], DFIRE2 score [47]. All of these scores are converted into the range of 0 and 1 as the input features for training the deep leaning networks. Occasionally, if a feature cannot be calculated for a model due to the failure of a tool, its value is set to 0.5.

The remaining 7 input features are generated from the physio-chemical properties of a protein model. These features are calculated from a structural model and its protein sequence [34], which include: secondary structure similarity (SS) score, solvent accessibility similarity (SA) score, secondary structure penalty (SP) score, Euclidean compact (EC) score, Surface (SU) score, exposed mass (EM) score, exposed surface (ES) score.

A summary table of all features used for benchmarking DeepQA is given in **Table 1**.

**Deep belief network architectures and training procedure**

Our in-house deep belief network framework [40] is used to train deep learning models for protein model quality assessment. As is shown in **Figure 1**, in this framework, a two-layer Restricted Boltzmann Machines (RBMs) form the hidden layers of the deep learning networks, and one layer of logistic regression node is added at the top to output a real value between 0 and 1 as predicted quality score. The weights of RBMs are initialized by unsupervised learning called pre-training. The pre-train process is carried out by the 'contrastive divergence' algorithm to adjust the weight in the RBM networks [48]. The mean square error is considered as cost function in the process of standard error backward propagation. The final deep belief architecture is fine-tuned and optimized based on Broyden-Fletcher-Goldfarh-Shanno(BFGS) optimization [49]. We divide the training data equally into five sets, and a five-fold cross validation is used to train and validate DeepQA. Five parameters of DeepQA are adjusted during the training procedure. The five parameters are total number of nodes at the first hidden layer (N1), total number of nodes at the second hidden layer (N2), learning rate $\varepsilon$ (default 0.001), weight cost $\omega$ (default 0.07), and momentum $\nu$ (default from 0.5 to 0.9). The last three parameters are used for training the RBMs. The average of Mean Absolute Error (MAE) is calculated for each round of five-fold cross validation to estimate the model accuracy. MAE is the absolute difference of predicted value and real value.

**Model accuracy evaluation metrics**

We evaluate the accuracy of DeepQA on 84 protein targets on both stage 1 and stage 2 models of the 11[th] community-wide experiment on the Critical Assessment of Techniques for Protein Structure Prediction (CASP11), which are available in the CASP official website (http://www.predictioncenter.org/casp11/index.cgi).

The real GDT-TS score of each protein model is calculated against the native structure by TM-score [50]. Second, all feature scores are calculated for each protein model. The trained DeepQA is used to predict the quality score of a model based on its feature scores.

To evaluate the performance of QA method, we use the following metrics: average per-target loss which is the difference of GDT-TS score of the top 1 model selected by a QA method and that of the best model in the model pool, average per-target correlation which is the Pearson's correlation between all models' real GDT-TS scores and its predicted scores, the summation of real TM-score and RMSD scores of the top models selected by a QA method, and the summation of real TM-score and RMSD scores of the best of top 5 models selected by QA methods.

To evaluate the performance of QA methods on *ab initio* models, we calculated the average per-target TM-score and RMSD for the selected best model, and also the selected best of top 5 models by QA methods.

## Results and Discussion

### Comparison of Deep learning with support vector ma-chines and neural networks

We train the deep learning and two other most widely used machine learning techniques (Support Vector Machine and Neural Network) separately on our training datasets and compare their performance using five-fold cross-validation protocol. SVMlight [7] is used to train the support vector machine, and the tool Weka [51] is used to train the neural networks. The RBF kernel function is used for support vector machine, and the following three parameters are adjusted: C for the trade-off between training error and margin, $\varepsilon$ for the epsilon width of tube for regression, and parameter gamma for RBF kernel. We randomly select 7500 data points from the whole datasets to form a small dataset to estimate these parameters of support vector machine to speed up the training process. Based on the cross validation result on this selected small dataset, C is set to 60, Ɛ to 0.19, gamma to 0.95. For the neural network, we adjust the following three parameters: the number of hidden nodes in the first layer (from 5 to 40), the number of hidden nodes in the second layer (from 5 to 40), and the learning rate (from 0.01 to 0.4). Based on the cross validation result on the entire datasets, we set the number of hidden nodes as 40 and 30 for the first and second layer respectively, and the learning rate is set to be 0.3. For the deep belief network, we test the number of hidden nodes in the first and second layer of RBMs from 5 to 40 respectively, learning rate Ɛ from 0.0001 to 0.01, weight cost ω from 0.001 to 0.7, and momentum ν from 0.5 to 0.9. Based on the MAE of cross validation result, we find the following parameters with good performance: the number of hidden nodes in the first and second layer of RBMs is set to 20 and 10 respectively, learning rate to 0.0001, weight cost to 0.007, and momentum from 0.5 to 0.9.

The correlation and loss on both stage 1 and stage 2 models of CASP11 datasets are calculated for these three methods, and the results are shown in **Table 2**. Deep belief network has the best average per-target correlation on both stage 1 and stage 2. The loss of DeepQA is also lower than or equal to the other two methods. The results suggest that deep belief network is a good choice for protein quality assessment problem.

**Comparison of DeepQA with other single-model QA methods on CASP11**

In order to reduce the model complexity and improve accuracy, we do a further analysis by selecting good features out of all these 16 features for our method DeepQA. First of all, we fix a set of parameters with good performance on all 16 features (e.g, the number of nodes in the first and second hidden layer is set to 20 and 10 respectively), and then train the Deep Belief Network for different combination of all these 16 features. Based on the MAE of these models in the training datasets, we use the following features which has relatively good performance and also low model complexity as the final features of DeepQA: Surface score, Dope score, GOAP score, OPUS score, RWplus score, Modelevaluator score, Secondary structure penalty score, Euclidean compact score, and Qprob score.

We evaluate the DeepQA on CASP11 datasets, and compare it with other single-model QA methods participating in CASP11. We use the standard evaluation metrics – average per-target correlation and aver-age per-target loss based on GDT-TS score to evaluate the performance of each method (see the results in **Table 3**). On stage 1 of CASP11, the average per-target correlation of DeepQA is 0.64, which is the same as the ProQ2 and better than Qprob. The average per-target loss of DeepQA is 0.09, same as ProQ2 and ProQ2-refine, and better than other single-model QA methods. On stage 2 models of CASP11, DeepQA has the highest per-target average correlation. Its per-target average loss is the same as ProQ2, and better than all other QA methods. Overall, the results demonstrate that DeepQA has achieved the state-of-the-art performance.

In order to evaluate how DeepQA aids the protein tertiary structure prediction methods in model selection, we apply DeepQA to select models in the stage 2 dataset of CASP11 submitted by top performing protein tertiary structure prediction methods. For most cases, DeepQA helps the prtein tertiary structure prediction methods to improve the quality of the top selected model. For example, DeepQA improves overall Z-score for Zhang-Server by 6.39, BAKER-ROSETTASERVER by 16.34, and RaptorX by 6.66. The result of applying DeepQA on 10 top performing protein tertiary structure prediction methods is shown at **Supplementary Table S1**.

**Case study of DeepQA on *ab initio* datasets**

In order to assess the ability of DeepQA in evaluating *ab initio* models, we evaluate it on 24 *ab initio* targets with more than 20,000 models generated by UniCon3D. **Table 4** shows the average per-target TM-score and RMSD for the top one model and best of top 5 models selected by DeepQA, ProQ2, and two energy scores (i.e., Dope and RWplus), respectively. The result shows DeepQA achieves good performance in terms of TM-score and RMSD compared with ProQ2 and two top-performing energy scores. The TM-score difference of best of top 5 models between DeepQA and ProQ2 is significant. **Supplementary Tables S2 and S3** show the per-target TM-score and RMSD of DeepQA and ProQ2 on this *ab initio* datasets.

## Conclusions

In this paper, we develop a single-model QA method (DeepQA) based on deep belief network. It performs better than support vector machines and neural networks, and achieve the state-of-the-art performance in comparison with other established QA methods. DeepQA is also useful for ranking *ab initio* protein models. And DeepQA could be further improved by incorporating more relevant features and training on larger datasets.

## Availability and requirements

**Project name:** DeepQA

**Project homepage:** http://cactus.rnet.missouri.edu/DeepQA/

**Operating Systems:** Linux

**Programming language:** Perl

# Competing interests

# Authors' contributions

# Acknowledgements

This work is partially supported by NIH R01 (R01GM093123) grant to JC.

# References

1. Jacobson M, Sali A: **Comparative protein structure modeling and its applications to drug discovery**. *Annu Rep Med Chem* 2004, **39**(85):259-274.
2. Li J, Cao R, Cheng J: **A large-scale conformation sampling and evaluation server for protein tertiary structure prediction and its assessment in CASP11**. *BMC bioinformatics* 2015, **16**(1):337.
3. Cao R, Cheng J: **Integrated protein function prediction by mining function associations, sequences, and protein–protein and gene–gene interaction networks**. *Methods* 2016, **93**:84-91.
4. Cao R, Bhattacharya D, Adhikari B, Li J, Cheng J: **Large-scale model quality assessment for improving protein tertiary structure prediction**. *Bioinformatics* 2015, **31**(12):i116-i123.
5. Cao R, Jo T, Cheng J: **Evaluation of Protein Structural Models Using Random Forests**. *arXiv preprint arXiv:160204277* 2016.
6. Li J, Bhattacharya D, Cao R, Adhikari B, Deng X, Eickholt J, Cheng J: **The MULTICOM protein tertiary structure prediction system**. *Protein Structure Prediction* 2014, **1137**:29-41.
7. Joachims T: **Optimizing search engines using clickthrough data**. In: *Proceedings of the eighth ACM SIGKDD international conference on Knowledge discovery and data mining: 2002*. ACM: 133-142.
8. Simons KT, Kooperberg C, Huang E, Baker D: **Assembly of protein tertiary structures from fragments with similar local sequences using simulated annealing and Bayesian scoring functions**. *Journal of molecular biology* 1997, **268**(1):209-225.
9. Page R: **TreeView: an application to display phylogenetic trees on personal computer**. *Comp Appl Biol Sci* 1996, **12**:357 - 358.

10. Wang C, Zhang H, Zheng W-M, Xu D, Zhu J, Wang B, Ning K, Sun S, Li SC, Bu D: **FALCON@ home: a high-throughput protein structure prediction server based on remote homologue recognition**. *Bioinformatics* 2015:btv581.
11. Li SC, Bu D, Xu J, Li M: **Fragment‐HMM: A new approach to protein structure prediction**. *Protein Science* 2008, **17**(11):1925-1934.
12. Zhang Y: **I-TASSER server for protein 3D structure prediction**. *BMC bioinformatics* 2008, **9**(1):40.
13. Zhang J, Wang Q, Barz B, He Z, Kosztin I, Shang Y, Xu D: **MUFOLD: a new solution for protein 3D structure prediction**. *Proteins* 2010, **78**(5):1137 - 1152.
14. Peng J, Xu J: **RaptorX: exploiting structure information for protein alignments by statistical inference**. *Proteins* 2011, **79**(S10):161 - 171.
15. Liaw A, Wiener M: **Classification and regression by randomForest**. *R news* 2002, **2**(3):18-22.
16. Zhang W, Chen J, Yang Y, Tang Y, Shang J, Shen B: **A practical comparison of de novo genome assembly software tools for next-generation sequencing technologies**. *PLoS One* 2011, **6**(3):e17915.
17. Bhattacharya D, Cheng J: **De novo protein conformational sampling using a probabilistic graphical model**. *Scientific Reports* 2015, **5**.
18. Liu T, Wang Y, Eickholt J, Wang Z: **Benchmarking Deep Networks for Predicting Residue-Specific Quality of Individual Protein Models in CASP11**. *Scientific Reports* 2016, **6**:19301.
19. Bhattacharya D, Cao R, Cheng J: **UniCon3D: de novo protein structure prediction using united-residue conformational search via stepwise, probabilistic sampling**. *Bioinformatics* 2016:btw316.
20. Adhikari B, Bhattacharya D, Cao R, Cheng J: **CONFOLD: Residue‐residue contact‐guided ab initio protein folding**. *Proteins: Structure, Function, and Bioinformatics* 2015, **83**(8):1436-1449.
21. McGuffin L: **The ModFOLD server for the quality assessment of protein structural models**. *Bioinformatics* 2008, **24**(4):586 - 587.
22. Wang Q, Vantasin K, Xu D, Shang Y: **MUFOLD-WQA: a new selective consensus method for quality assessment in protein structure prediction**. *Proteins* 2011, **79**(SupplementS10):185 - 195.
23. McGuffin L, Roche D: **Rapid model quality assessment for protein structure predictions using the comparison of multiple models without structural alignments**. *Bioinformatics* 2010, **26**(2):182 - 188.
24. Cao R, Wang Z, Cheng J: **Designing and evaluating the MULTICOM protein local and global model quality prediction methods in the CASP10 experiment**. *BMC structural biology* 2014, **14**(1):13.
25. Cao R, Wang Z, Wang Y, Cheng J: **SMOQ: a tool for predicting the absolute residue-specific quality of a single protein model with support vector machines**. *BMC bioinformatics* 2014, **15**(1):120.
26. Rykunov D, Fiser A: **Effects of amino acid composition, finite size of proteins, and sparse statistics on distance‐dependent statistical pair potentials**. *Proteins: Structure, Function, and Bioinformatics* 2007, **67**(3):559-568.

27. Zhang J, Zhang Y: **A novel side-chain orientation dependent potential derived from random-walk reference state for protein fold selection and structure prediction**. *PLoS One* 2010, **5**(10):e15386.
28. Wang Z, Tegge AN, Cheng J: **Evaluating the absolute quality of a single protein model using structural features and support vector machines**. *Proteins* 2009, **75**(3):638-647.
29. Shen My, Sali A: **Statistical potential for assessment and prediction of protein structures**. *Protein Science* 2006, **15**(11):2507-2524.
30. Cao R, Cheng J: **Protein single-model quality assessment by feature-based probability density functions**. *Scientific Reports* 2016, **6**:23990.
31. Kalman M, Ben-Tal N: **Quality assessment of protein model-structures using evolutionary conservation**. *Bioinformatics* 2010, **26**(10):1299 - 1307.
32. Liithy R, Bowie J, Eisenberg D: **Assessment of protein models with three-dimensional profiles**. *Nature* 1992, **356**:83 - 85.
33. Ray A, Lindahl E, Wallner B: **Improved model quality assessment using ProQ2**. *BMC bioinformatics* 2012, **13**(1):224.
34. Mishra A, Rao S, Mittal A, Jayaram B: **Capturing native/native like structures with a physico-chemical metric (pcSM) in protein folding**. *Biochimica et Biophysica Acta (BBA)-Proteins and Proteomics* 2013, **1834**(8):1520-1531.
35. Benkert P, Biasini M, Schwede T: **Toward the estimation of the absolute quality of individual protein structure models**. *Bioinformatics* 2011, **27**(3):343-350.
36. Uziela K, Wallner B: **ProQ2: Estimation of Model Accuracy Implemented in Rosetta**. *Bioinformatics* 2016:btv767.
37. LeCun Y, Bengio Y, Hinton G: **Deep learning**. *Nature* 2015, **521**(7553):436-444.
38. Zou WY, Wang X, Sun M, Lin Y: **Generic object detection with dense neural patterns and regionlets**. *arXiv preprint arXiv:14044316* 2014.
39. Silver D, Huang A, Maddison CJ, Guez A, Sifre L, van den Driessche G, Schrittwieser J, Antonoglou I, Panneershelvam V, Lanctot M: **Mastering the game of Go with deep neural networks and tree search**. *Nature* 2016, **529**(7587):484-489.
40. Eickholt J, Cheng J: **Predicting protein residue–residue contacts using deep networks and boosting**. *Bioinformatics* 2012, **28**(23):3066-3072.
41. Deng H, Jia Y, Zhang Y: **3DRobot: automated generation of diverse and well-packed protein structure decoys**. *Bioinformatics* 2015:btv601.
42. Wang G, Dunbrack RL: **PISCES: a protein sequence culling server**. *Bioinformatics* 2003, **19**(12):1589-1591.
43. Simons K, Kooperberg C, Huang E, Baker D: **Assembly of protein tertiary structures from fragments with similar local sequences using simulated annealing and Bayesian scoring functions**. *J Mol Biol* 1997, **268**(1):209 - 225.
44. John B, Sali A: **Comparative protein structure modeling by iterative alignment, model building and model assessment**. *Nucleic acids research* 2003, **31**(14):3982-3992.
45. Zhou H, Skolnick J: **GOAP: a generalized orientation-dependent, all-atom statistical potential for protein structure prediction**. *Biophysical journal* 2011, **101**(8):2043-2052.
46. Wu Y, Lu M, Chen M, Li J, Ma J: **OPUS‐Ca: A knowledge‐based potential function requiring only Cα positions**. *Protein Science* 2007, **16**(7):1449-1463.

47. Yang Y, Zhou Y: **Specific interactions for ab initio folding of protein terminal regions with secondary structures**. *Proteins: Structure, Function, and Bioinformatics* 2008, **72**(2):793-803.
48. Hinton GE: **Training products of experts by minimizing contrastive divergence**. *Neural computation* 2002, **14**(8):1771-1800.
49. Nawi NM, Ransing MR, Ransing RS: **An improved learning algorithm based on the Broyden-Fletcher-Goldfarb-Shanno (BFGS) method for back propagation neural networks**. In: *Intelligent Systems Design and Applications, 2006 ISDA'06 Sixth International Conference on: 2006*. IEEE: 152-157.
50. Zhang Y, Skolnick J: **Scoring function for automated assessment of protein structure template quality**. *Proteins: Structure, Function, and Bioinformatics* 2004, **57**(4):702-710.
51. Hall M, Frank E, Holmes G, Pfahringer B, Reutemann P, Witten IH: **The WEKA data mining software: an update**. *ACM SIGKDD explorations newsletter* 2009, **11**(1):10-18.
52. Cheng J, Randall AZ, Sweredoski MJ, Baldi P: **SCRATCH: a protein structure and structural feature prediction server**. *Nucleic Acids Research* 2005, **33**(suppl 2):W72-W76.
53. Kabsch W, Sander C: **Dictionary of protein secondary structure: pattern recognition of hydrogen‐bonded and geometrical features**. *Biopolymers* 1983, **22**(12):2577-2637.
54. Faraggi E, Zhang T, Yang Y, Kurgan L, Zhou Y: **SPINE X: improving protein secondary structure prediction by multistep learning coupled with prediction of solvent accessible surface area and backbone torsion angles**. *Journal of computational chemistry* 2012, **33**(3):259-267.

## Tables

**Table 1.** 16 features for benchmarking DeepQA.

| Feature Name | Feature descriptions |
| --- | --- |
| (1). Surface score (SU) | The total area of exposed nonpolar residues divided by the total area of all residues |
| (2). Exposed mass score (EM) | The percentage of mass for exposed area, equal to the total mass of exposed area divided by the total mass of all area |
| (3). Exposed surface score | The total exposed area divided by the total area |

| (ES) | |
|---|---|
| (4). Solvent accessibility score (SA) | The difference of solvent accessibility predicted by SSpro4[52] from the protein sequence and those of a model parsed by DSSP [53] |
| (5). RF_CB_SRS_OD score[26] | A novel distance dependent residue-level potential energy score. |
| (6). DFIRE2 score [47] | A distance-scaled all atom energy score. |
| (7). Dope score [29] | A new statistical potential discrete optimized protein energy score. |
| (8). GOAP score [45] | A generalized orientation-dependent, all-atom statistical potential score. |
| (9). OPUS score [46] | A knowledge-based potential score. |
| (10). ProQ2 score [36] | A single-model quality assessment method by machine learning techniques. |
| (11). RWplus score [27] | A new energy score using pairwise distance-dependent atomic statistical potential function and side-chain orientation-dependent energy term |
| (12). ModelEvaluator score [28] | A single-model quality assessment score based on structural features using support vector machine. |
| (13). Secondary structure similarity score (SS) | The difference of secondary structure information predicted by Spine X [54] from a protein sequence and those of a model parsed by DSSP [53] |
| (14). Secondary structure | Calculated from the predicted secondary structure alpha- |

| | | | | | |
|---|---|---|---|---|---|
| penalty score (SP) | helix and beta-sheet matching with the one parsed by DSSP. | | | | |
| (15). Euclidean compact score (EC) | The pairwise Euclidean distance of all residues divided by the maximum Euclidean distance (3.8) of all residues. | | | | |
| (16). Qprob [30] | A single-model quality assessment score that utilizes 11 structural and physicochemical features by feature-based probability density functions. | | | | |

**Table 2**. The accuracy of Deep Belief Network, Support Vector Machines, and Neural Networks in terms of MAE based on cross validation of training datasets, the average per-target correlation, and loss on stage 1 and stage 2 of CASP11 datasets for all three difference techniques.

| | MAE based on cross validation | Corr. on stage 1 | Loss on stage 1 | Corr. on stage 2 | Loss on stage 2 |
|---|---|---|---|---|---|
| Deep Belief Network | 0.08 | 0.63 | 0.09 | 0.34 | 0.06 |
| Support Vector Machine | 0.12 | 0.58 | 0.10 | 0.32 | 0.07 |
| Neural Network | 0.08 | 0.51 | 0.12 | 0.25 | 0.07 |
| Mean | 0.09 | 0.57 | 0.10 | 0.30 | 0.07 |

**Table 3**. Average per-target correlation and loss for DeepQA and other top performing single-model QA methods on CASP11. The table is ranked based on the average per-target loss on stage 2 of CASP11.

| QA methods | Corr. on stage 1 | Loss on stage 1 | Corr. on stage 2 | Loss on stage 2 |
|---|---|---|---|---|
| DeepQA | 0.64 | 0.09 | 0.42 | 0.06 |
| ProQ2 | 0.64 | 0.09 | 0.37 | 0.06 |
| Qprob | 0.63 | 0.10 | 0.38 | 0.07 |
| VoroMQA | 0.56 | 0.11 | 0.40 | 0.07 |
| ProQ2-refine | 0.65 | 0.09 | 0.37 | 0.07 |
| Wang_SVM | 0.66 | 0.11 | 0.36 | 0.09 |
| raghavagps-qaspro | 0.35 | 0.16 | 0.22 | 0.09 |
| Wang_deep_2 | 0.63 | 0.12 | 0.31 | 0.09 |
| Wang_deep_1 | 0.61 | 0.13 | 0.30 | 0.09 |
| Wang_deep_3 | 0.63 | 0.12 | 0.30 | 0.09 |
| FUSION | 0.10 | 0.15 | 0.05 | 0.11 |
| Mean | 0.55 | 0.12 | 0.32 | 0.08 |

**Table 4**. Model selection ability on *ab initio* datasets for DeepQA, ProQ2, Dope2, and RWplus score

| QA methods | TM-score on top 1 model | RMSD on top 1 model | TM-score on best of top 5 | RMSD on best of top 5 |
|---|---|---|---|---|
| DeepQA | 0.23 | 19.01 | 0.26 | 17.14 |
| ProQ2 | 0.22 | 19.73 | 0.25 | 17.93 |
| Dope | 0.22 | 19.55 | 0.24 | 18.10 |
| RWplus | 0.22 | 19.68 | 0.25 | 17.38 |
| Mean | 0.22 | 19.49 | 0.25 | 17.64 |

## Figures

**Figure 1**. The Deep Belief Network architecture for DeepQA.

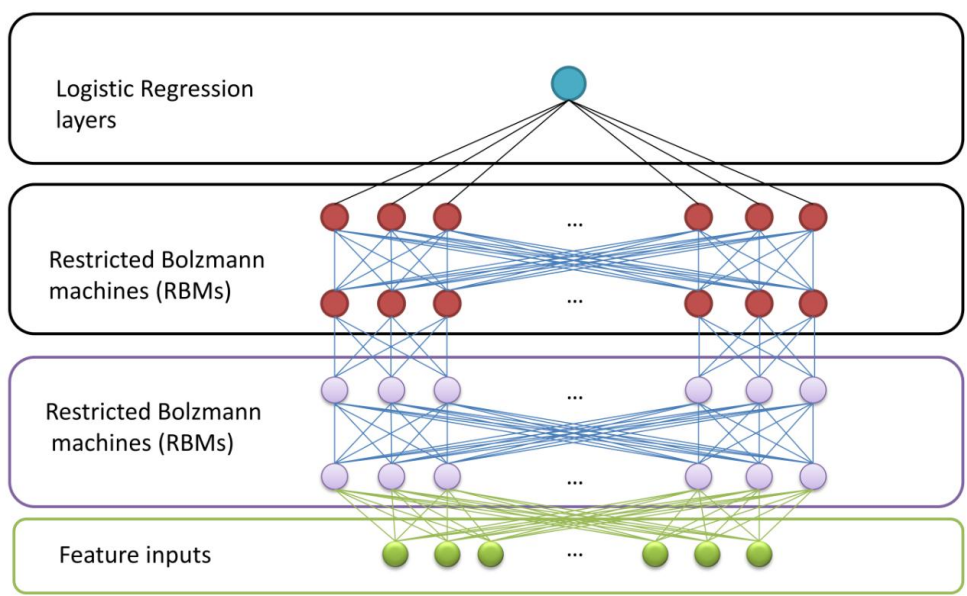